\documentclass[journal]{IEEEtran}
\usepackage{amsmath,amsfonts}
\usepackage{algorithmic}
\usepackage{algorithm}
\usepackage{array}
\usepackage[caption=false,font=normalsize,labelfont=sf,textfont=sf]{subfig}
\usepackage{textcomp}
\usepackage{stfloats}
\usepackage{url}
\usepackage{verbatim}
\usepackage{graphicx}
\usepackage{multirow}
\usepackage{amssymb}
\usepackage{xcolor}
\usepackage[capitalize]{cleveref}
\crefname{section}{Sec.}{Secs.}
\Crefname{section}{Section}{Sections}
\Crefname{table}{Table}{Tables}
\crefname{table}{Tab.}{Tabs.}
\crefname{equation}{Eq.}{Eqs.}

\usepackage{cite}
\begin{document}
\title{Leveraging Diffusion Model and Image Foundation Model for Improved Correspondence Matching in Coronary Angiography}
\author{Lin Zhao, Xin Yu, Yikang Liu, Xiao Chen, Eric Z. Chen, Terrence Chen, Shanhui Sun \thanks{L. Zhao, Y. Liu, X. Chen, E. Chen,  T. Chen and S. Sun are with United Imaging Intelligence, 65 Blue Sky Drive, Burlington, MA 01803, USA. (e-mail: shanhui.sun@uii-ai.com) } \thanks{X. Yu is with Department of Computer Science, Vanderbilt University, Nashville, TN 37212, USA}
\thanks{Contribution from X. Yu was carried out during her internship at United Imaging Intelligence, Burlington, MA, USA.}}

\maketitle

\begin{abstract}
Accurate correspondence matching in coronary angiography images is crucial for reconstructing 3D coronary artery structures, which is essential for precise diagnosis and treatment planning of coronary artery disease (CAD). Traditional matching methods for natural images often fail to generalize to X-ray images due to inherent differences such as lack of texture, lower contrast, and overlapping structures, compounded by insufficient training data. To address these challenges, we propose a novel pipeline that generates realistic paired coronary angiography images using a diffusion model conditioned on 2D projections of 3D reconstructed meshes from Coronary Computed Tomography Angiography (CCTA), providing high-quality synthetic data for training. Additionally, we employ large-scale image foundation models to guide feature aggregation, enhancing correspondence matching accuracy by focusing on semantically relevant regions and keypoints. Our approach demonstrates superior matching performance on synthetic datasets and effectively generalizes to real-world datasets, offering a practical solution for this task. Furthermore, our work investigates the efficacy of different foundation models in correspondence matching, providing novel insights into leveraging advanced image foundation models for medical imaging applications.
\end{abstract}


\section{Introduction}
\label{sec:introduction}
\IEEEPARstart{C}{oronary} artery disease (CAD), characterized by the narrowing or blockage of the coronary arteries, is the leading cause of death in the U.S. and worldwide\cite{go2014heart,townsend2016cardiovascular}. Catheter-based invasive coronary angiography (ICA) is one of the most commonly used imaging approaches for assessing and diagnosing CAD, and it remains the gold standard for clinical decision-making and intervention planning\cite{de2008fractional}. However, X-ray coronary angiography has fundamental limitations since it projects the 3D structure of contrast-filled coronary arteries into 2D X-ray images \cite{green2004three,katritsis1991limitations}, leading to a significant loss of 3D information. The resulting 2D X-ray images thus suffer from visual ambiguities such as vessel overlap, foreshortening, tortuosity, and eccentricity, which can result in inaccurate estimation of stenosis severity and even incorrect selection of stent size\cite{green2004three}.

\begin{figure}[t]
\begin{center}
\includegraphics[width=1\linewidth]{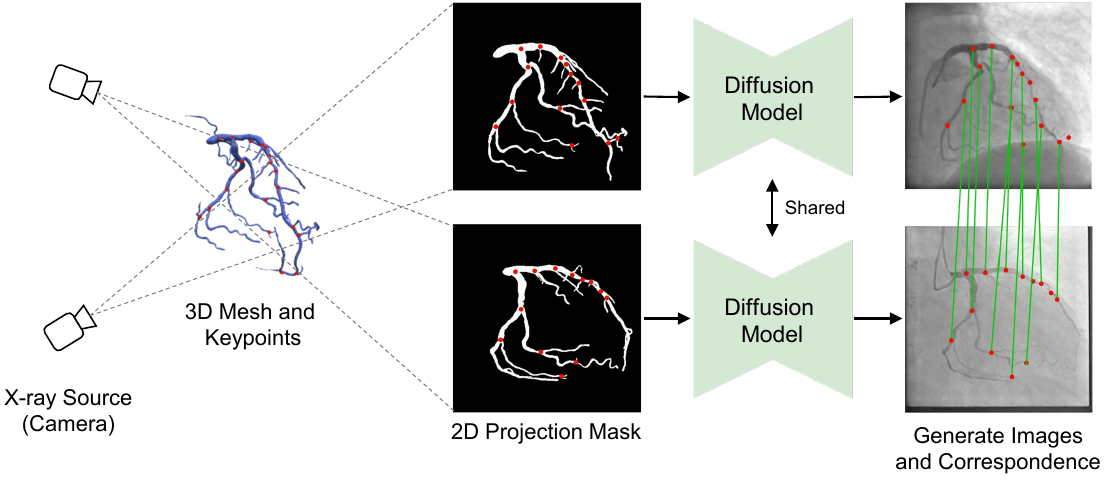}
\end{center}
\caption{Synthetic data generation. The mesh reconstructed from CCTA is projected as 2D masks from different angles. X-ray images are generated by a diffusion model conditioned on these masks. Correspondence of 2D keypoints is established by linking the corresponding points in 3D space.}
\label{fig:fig1}
\end{figure}

To overcome aforementioned limitations, coronary angiographies are typically acquired from multiple angles \cite{gollapudi2007utility,green2004three}, providing more information that allows for the reconstruction of a 3D representation of the coronary arteries from the 2D X-ray images\cite{yang2009novel,bourantas2005method,vukicevic2018three}. However, this reconstruction process presents significant challenges due to its nature as an ill-posed inverse problem\cite{ccimen2016reconstruction}. One of the main difficulties arises from respiratory and cardiac motions \cite{baka2015respiratory,blondel2006reconstruction}, which cause the coronary arteries to move during imaging. Typically, the 2D images from multiple angles are acquired by moving the C-arm of angiography system, which are not acquired simultaneously but at different times. This results in 2D images that capture varying respiratory and cardiac motions even within the same phase of cardiac cycle, leading to discrepancies in the relative positions of the coronary arteries. Consequently, accurate 3D reconstruction requires precise calibration based on correspondence matching between these 2D images to compensate for these movements and ensure the correct alignment and positioning of the coronary arteries.

Despite extensive research in computer vision filed on establishing correspondences in 2D images\cite{sarlin2020superglue,lindenberger2023lightglue}, correspondence matching for coronary angiography images presents unique domain-specific challenges. These challenges arise from the fundamental differences between X-ray and natural images. X-ray images are generated by the transmission of X-rays through a subject, capturing internal structures based on varying absorption levels. This process results in images that lack surface texture, distinctive patterns, and often have lower contrast in some areas. In contrast, natural images, formed by reflected light, contain rich textures, patterns, and distinctive features that facilitate easier matching. Overlapping structures in X-ray images further complicate accurate correspondence matching by causing ambiguities, especially in complex, tree-like structures such as coronary arteries. Usually, it is difficult to discern which artery is being imaged in overlapping regions, or a bifurcation points in 2D is real bifurcation in 3D. Thus, the matching methods perform well in natural image may not generalize well to coronary angiography images.

Additionally, adapting and finetuning correspondence matching algorithms for coronary angiography images are also hindered by the lack of enough training samples. Acquiring angiographies is inherently challenging, especially in the required paired format. For successful 3D reconstruction, the two views must be captured from sufficiently different angles\cite{xu2017diagnostic}, further limiting the availability of suitable data. Obtaining accurate annotations also poses a considerable challenge. Unlike natural images, where depth maps can be relatively easily obtained\cite{li2018megadepth}, angiography images lack straightforward depth information. Manual annotation is also problematic because these images lack distinctive features, making it difficult even for human annotators. Although bifurcations might serve as distinct landmarks, determining their precise positions in 2D that can accurately reproject to 3D is challenging due to discrepancies arising from parallax and the inherent thickness of the vessels. Furthermore, relying solely on bifurcations is insufficient, particularly for the right coronary artery (RCA), which has fewer bifurcations and may not provide enough data points to meet calibration needs. 

In this paper, we propose a novel pipeline to tackle the challenges of establishing accurate correspondences in coronary angiography images. Our approach begins with developing a novel data engine that generates paired 2D images using diffusion model (\Cref{fig:fig1}). Specifically, the mesh reconstructed from 3D Coronary Computed Tomography Angiography (CCTA) is projected as 2D masks from different angles. We then train a diffusion model, conditioned on these masks, to generate 2D X-ray images, as illustrated in \Cref{fig:fig1}). Additionally, the centerline points of the 3D mesh are projected onto the 2D images, establishing ground truth correspondences not only for landmarks such as bifurcations but also for indistinct regions. To enhance matching results, we adapt image matching models by leveraging large image foundation models. Recent advancements, such as DINOv2\cite{oquab2023dinov2}, have demonstrated the ability to capture image semantics. This semantic information can be crucial in coronary angiography image matching, guiding the matching process to exclude unrelated points and focus on semantically similar points.

Our method is evaluated on both synthetic and real datasets. Guided by image foundation models, the proposed method trained on generated data demonstrates superior matching performance on the synthetic dataset and generalizes well to real datasets. We also explore the efficacy of various foundation models in guiding angiography image matching. Our contributions in this work are summarized as follows:

\begin{itemize}
    \item We propose a novel data engine to generate realistic paired coronary angiography images based on the projection of 3D meshes and a diffusion model.
    \item We employ large-scale image foundation models to guide feature aggregation for correspondence matching, significantly improving the model's performance and generalizability from synthetic to real data.
    \item We explore the efficacy of different foundation models in enhancing the generalizability of matching methods, providing new insights into their effectiveness.
\end{itemize}

\section{Related Works}

\subsection{Correspondence Matching}

Correspondence matching in natural images involves identifying and aligning similar features across different images, which can be roughly categorized into sparse \cite{sarlin2020superglue,lindenberger2023lightglue,jiang2024omniglue} and dense methods\cite{liu2017deep,sun2021loftr}. Sparse methods focus on detecting and matching distinct key points or features. Recent advancements like SuperGlue\cite{sarlin2020superglue} and LightGlue\cite{lindenberger2023lightglue} have significantly improved the accuracy and robustness of sparse matching by leveraging deep learning techniques to enhance feature matching. Dense methods, which aim to establish correspondences for every pixel or region, provide a more comprehensive alignment. Methods such as LoFTR (Local Feature TRansformer)\cite{sun2021loftr} utilize transformer architectures to achieve high-precision dense matching.

In coronary angiography, correspondence matching is particularly challenging due to the nature of X-ray images and the complexity of vascular structures. Traditional methods often involve manual intervention and are not fully automated \cite{cong2013energy,cong2015quantitative,yang2014external,kunio2018vessel}. A recent approach, following the scheme of COTR (Correspondence Transformer) \cite{jiang2021cotr}, formulates the problem of correspondence matching as a regression problem to establish dense correspondence between angiography images. However, this method does not generalize well to real images, often resulting in corresponding points that are outside the vessels, making it impractical for real-world applications. Our method is based on feature matching of the centerline points, which provides a strong constraint on the position of matched points. Our matching process also benefits from the semantic understanding provided by foundation models like DINOv2 \cite{oquab2023dinov2}, which guide the matching process by focusing on semantically relevant regions and points, enabling more precise matching and pose estimation.

\subsection{Diffusion Model}
Diffusion models, such as Denoising Diffusion Probabilistic Models (DDPM) \cite{ho2020denoising}, Score-based diffusion models \cite{song2020score}, and Denoising Diffusion Implicit Models (DDIM) \cite{song2020denoising}, have revolutionized image generation tasks by progressively adding noise to data and learning to reverse this process to generate new samples. These models have also been applied in various medical imaging applications, such as image segmentation \cite{wu2024medsegdiff}, data synthesis \cite{chen2024federated}, and image reconstruction \cite{qiu2024clinically,qiu2024spatiotemporal}. In coronary angiography, diffusion models can be crucial for generating realistic 2D images. In this paper, we employ the Improved Denoising Diffusion Probabilistic Model (IDDPM) \cite{nichol2021improved} to generate realistic X-ray images based on 2D mask images projected from 3D meshes. These high-quality synthetic images are essential for training our correspondence matching model, significantly improving matching accuracy and pose estimation.

\subsection{Image Foundation Models}
Recent advancements in large-scale image foundation models, such as Vision Transformers (ViT) \cite{dosovitskiy2020image}, Contrastive Language-Image Pretraining (CLIP) \cite{radford2021learning}, Masked Autoencoders (MAE) \cite{he2022masked}, and DINO (self-Distillation with No labels) \cite{caron2021emerging,oquab2023dinov2}, have significantly enhanced computer vision tasks. ViT employs transformer architecture to capture long-range dependencies in images, while CLIP learns visual concepts from natural language descriptions, showcasing impressive zero-shot learning capabilities. MAE utilizes self-supervised learning by predicting masked parts of images, and DINO focuses on creating consistent representations without labeled data. DINOv2, an improvement over DINO, excels in learning high-level semantic information. A recent study used DINOv2 features to guide feature matching \cite{jiang2024omniglue}, greatly improving model generalizability in zero-shot settings on natural images. This inspired us to employ image foundation models for medical imaging tasks like coronary angiography image matching. By leveraging the semantic understanding of foundation models, the matching process becomes more accurate, focusing on semantically relevant regions and keypoints, thereby improving overall performance.

\section{Methods}

\subsection{Synthetic Data Engine}

The main challenge in training or adapting a correspondence matching model for coronary angiography images is the scarcity of paired images and the unavailability of annotations. Our pipeline addresses this issue by developing a novel data engine. This engine projects the 3D meshes of coronary arteries, reconstructed from CCTA, as paired 2D masks with different projection angles. Conditioned on these masks, the paired 2D X-ray images are generated with a diffusion model.

Specifically, we reconstruct the 3D meshes of the left coronary artery (LCA) and right coronary artery (RCA) from CCTA segmentation. From these meshes, we extract the 3D centerline points and bifurcation points. The reconstructed 3D meshes are then rendered as 2D silhouettes, using the projection angles recommended in \cite{xu2017diagnostic}.  Using the same camera extrinsic and intrinsic parameters, we project the 3D centerline points and bifurcation points into 2D. Since the corresponding 3D points for each 2D point are already known, we can establish ground truth correspondence for each paired masks.

The generation of the 2D X-ray images is achieved using the Improved Denoising Diffusion Probabilistic Model (IDDPM) \cite{nichol2021improved}, which involves two Gaussian processes: forward and reverse diffusion. In the forward phase, an uncorrupted image $x_0$ (angiography image) undergoes $T$ diffusion steps, progressively incorporating noise $\epsilon$, resulting in a sequence of increasingly noisy images ${x_0, \ldots, x_T}$. This process, modeled as a Gaussian process, follows a specified noise schedule ${\alpha_0, \ldots, \alpha_T}$:

\begin{equation}
x_t \sim \mathcal{N}(\sqrt{\bar{\alpha_t}} x_0, (1 - \bar{\alpha_t}) \mathbf{I})
\end{equation}

\noindent where $\bar{\alpha_t}$ is the cumulative product of the noise coefficients up to time $t$, defined as $\bar{\alpha_t} = \prod_{i=1}^{t} \alpha_i$.

The reverse process utilizes a synthesis network, parameterized by $\theta$ and conditioned on a vessel mask $M$, aims to reconstruct the original image from its noise-distorted counterpart by approximating the noise $\epsilon$. This process can be mathematically formulated as:

\begin{equation}
x_{t-1} \sim \mathcal{N}(\mu_{\theta}(x_t, t, M), \sigma^2_{\theta}(x_t, t, M) \mathbf{I})
\end{equation}

\noindent where

\begin{equation}
\mu_{\theta}(x_t, t, M) = \frac{1}{\sqrt{\alpha_t}} \left( x_t - \frac{1 - \alpha_t}{\sqrt{1 - \bar{\alpha_t}}} \epsilon_{\theta}(x_t, t, M) \right)
\end{equation}

Here, $\sigma_{\theta}(x_t, t, M)$ is modulated by a learned variance parameter $v_{\theta}$:

\begin{align}
\sigma^2_{\theta}(x_t, t, M) = & \exp \left( v_{\theta} \log(1 - \alpha_t) \right. \notag \\
& \left. + (1 - v_{\theta}) \log \left( 1 - \frac{\prod_{i=1}^{t-1} \alpha_i}{\prod_{i=1}^{t} \alpha_i} (1 - \alpha_t) \right) \right)
\end{align}

The IDDPM model is pretrained on a large dataset of coronary angiography images and corresponding masks to optimize the parameters $\mu_{\theta}$ and $\sigma_{\theta}$, following the method in \cite{nichol2021improved}. During inference, the process begins with Gaussian noise and a given vessel mask. The model then iteratively refines this noisy through the reverse diffusion process. With the pretrained synthesis network, the model reconstructs the  images step-by-step, progressively reducing the noise and incorporating the structural information from the mask. This results in the generation of high-quality, realistic angiography images of coronary arteries.

\subsection{Correspondence Matching Framework}

\begin{figure*}[t]
\begin{center}
\includegraphics[width=1\linewidth]{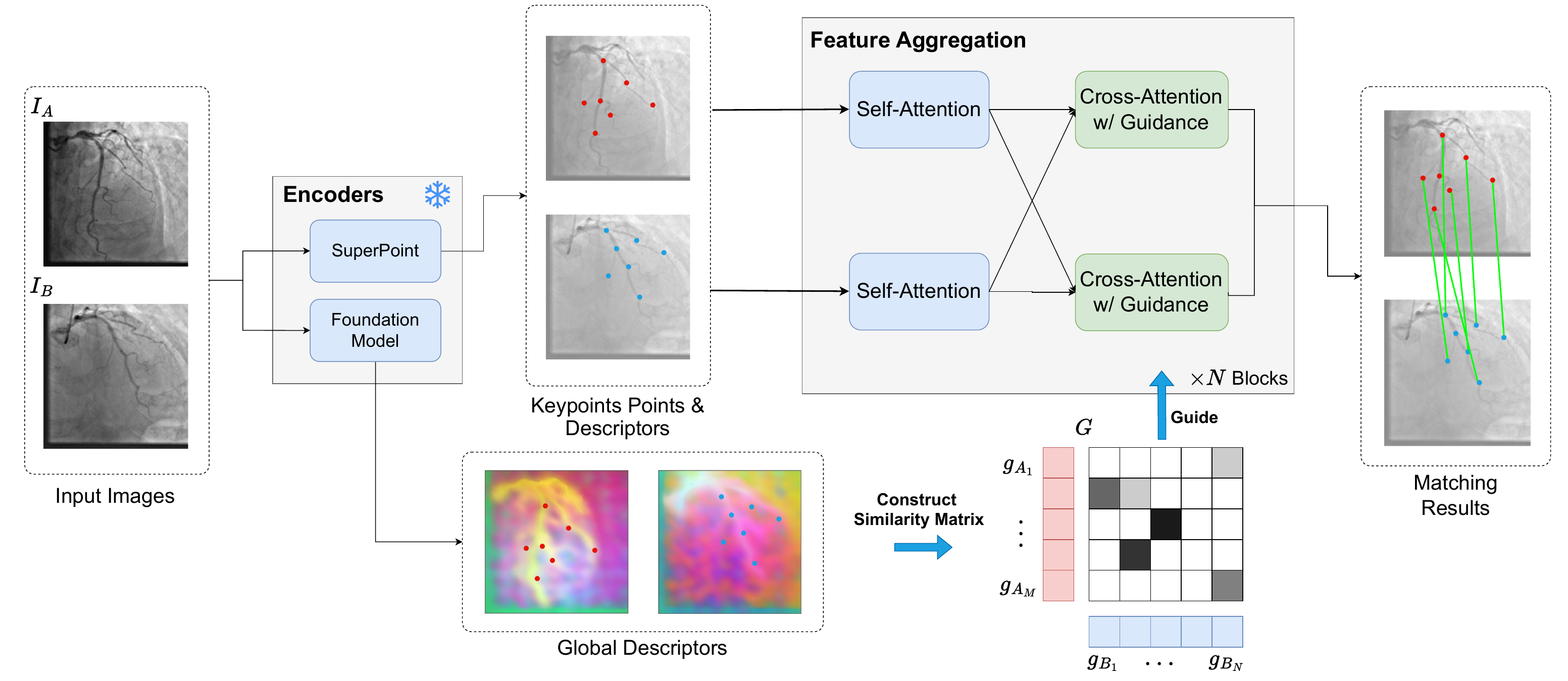}
\end{center}
\caption{The proposed matching framework. The paired images are first input into image encoders to extract local descriptors (from the SuperPoint model) and global descriptors (from the foundation model). Keypoints and local descriptors are fed into \(N\) consecutive feature aggregation blocks comprising self-attention and cross-attention layers. The global features are used to construct a similarity matrix and mask, guiding the cross-attention operation. This process filters out semantically irrelevant keypoints in feature aggregation, allowing the model to focus more on relevant regions and points.}
\label{fig:fig2}
\end{figure*}

We formulate the problem of correspondence matching as a keypoint-based feature matching problem. Given two input angiography images, denoted as $I_A$ and $I_B$, our task is to partially match the sets of keypoints $A: = \{A_1, \ldots, A_M\}$ for $I_A$ and $B: = \{B_1, \ldots, B_N\}$ for $I_B$, where $M$ and $N$ are the number of keypoints in $I_A$ and $I_B$, respectively. Each keypoint $A_i$ and $B_i$ is associated with a descriptor $d_{A_i}$ and $d_{B_i}$, respectively, where $d_{A_i}, d_{B_i} \in \mathbb{R}^d$. Our objective is to find a subset of pairs $(A_i, B_j)$ such that the corresponding descriptors are sufficiently similar.

Note that in our task, we do not adopt keypoints from detection methods such as SuperPoint \cite{detone2018superpoint}. Instead, the keypoints are the centerline points of the vessel derived from 2D vessel segmentation. This is based on the assumption that the cross-section of the vessel is either circular or oval. Consequently, points on the 2D centerline accurately represent the centerline of the vessel in 3D space and do not shift to the vessel's edge when viewed from different perspectives. We then sample the local descriptor $d$ for each keypoint by interpolating the feature maps of SuperPoint \cite{detone2018superpoint}.

With the keypoint descriptors, we enhance these descriptors by aggregating both spatial information and local features within and between images. We represent the local features of keypoints $A_i$ and $B_i$ as $x_{A_i}$ and $x_{B_i}$, which are initialized with the descriptors $x_{A_i} \leftarrow d_{A_i}$ and $x_{B_i} \leftarrow d_{B_i}$. Additionally, each keypoint is associated with a position embedding $p_{A_i}$ and $p_{B_i}$ which provide spatial context. We follow the methods in \cite{lindenberger2023lightglue} to normalize the locations of the keypoint and with a MLP to generate the final position embeddings. The local features are then updated through successive feature aggregation blocks. 

\subsubsection{Feature Aggregation Block}

As illustrated in \Cref{fig:fig2}, each feature aggregation block comprises one self-attention layer and one cross-attention layer. 

Self-attention allows each keypoint within an image to attend to other keypoints in the same image to incorporate related information. Here, for convenience, we disregard the distinction between keypoints in images $A$ and $B$ as the process is identical for both. For each keypoint $i$ in an image $I$, the current local features $x_i$ are transformed into query $q_i$, key $k_i$, and value $v_i$. The attention score $a_{ij}$ between keypoints $i$ and $j$ is computed as:

\begin{equation}
a_{ij} =\frac{(q_i+p_i) \cdot (k_j+p_j)}{\sqrt{d}},
\end{equation}

The message $m_{i_\text{self}}$ for keypoint $i$ is then computed as a weighted sum of the value vectors $v_j$ of all keypoints $j$ in the same image:

\begin{equation}
m_{i_\text{self}} = \sum_{j} \mathop\text{Softmax}_{\substack{k \in I}}\left( a_{ik} \right)_j v_j
\end{equation}

The local feature $x_i$ is then updated by integrating this message through an MLP with a residual connection:

\begin{equation}
x_i \leftarrow x_i + \text{MLP}([x_i \mid m_{i_\text{self}}]).
\end{equation}

Cross-attention enables keypoints in one image to attend to keypoints in another image, facilitating inter-image feature aggregation. For each keypoint $i$ in image $I$, the current local features $x_i$ are transformed into a query vector $q_i$, while the keypoints $j$ in the other image $S$ are transformed into key vectors $k_j$ and value vectors $v_j$. The attention score $a_{ij}$ between keypoint $i$ in image $I$ and keypoint $j$ in image $S$ is computed as:

\begin{equation}
a_{ij} = \frac{q_i \cdot k_j}{\sqrt{d}},
\end{equation}

The message $m_{i_\text{cross}}$ for keypoint $i$ is then computed as a weighted sum of the value vectors $v_j$ of all keypoints $j$ in the other image $S$:

\begin{equation}
m_{i_\text{cross}} = \sum_{j \in S} \mathop{\text{Softmax}}_{\substack{k \in S}} \left( a_{ik} \right)_j v_j
\end{equation}

Finally, the local feature $x_i$ is updated by integrating this cross-attention message through an MLP with a residual connection:

\begin{equation}
x_i \leftarrow x_i + \text{MLP}([x_i \mid m_{i_\text{cross}}]).
\end{equation}

\subsubsection{Correspondence Matching Layer}

With the updated local features $x_{A_i}$ and $x_{B_j}$ processed by the feature aggregation blocks, we compute the similarity matrix $S \in \mathbb{R}^{M \times N}$ between the points of both images:

\begin{equation}
S_{ij} = \text{f}(x_{A_i}) \cdot \text{f}(x_{B_j}) \quad \forall (i, j) \in A \times B,
\end{equation}

\noindent where $\text{f}(\cdot)$ is a learned linear transformation with bias. Following \cite{lindenberger2023lightglue}, we compute a matchability score for each point, indicating the likelihood that point $i$ has a corresponding point in the other image:

\begin{equation}
\sigma_i = \text{Sigmoid}(\text{f}(x_i)) \in [0, 1].
\end{equation}

The final soft partial assignment matrix $P$ is then computed as:

\begin{equation}
P_{ij} = \sigma_{A_i} \sigma_{B_j} \mathop{\text{Softmax}}_{\substack{k \in A}} (S_{kj})_i \mathop{\text{Softmax}}_{\substack{k \in B}} (S_{ik})_j
\end{equation}

We select pairs for which $P_{ij}$ is greater than a threshold $\tau$ and is the maximum element along both its row and column as the final matched pairs. The matching framework is optimized by minimizing the negative log-likelihood of the  assignment matrix $P$ follows the works \cite{lindenberger2023lightglue}.

\subsection{Matching Guided by Foundation Models}

Unlike natural images, angiography images lack the rich textures and distinctive patterns that facilitate straightforward feature matching. In angiography, different branches or bifurcations of vessels often appear very similar, making it difficult to match the correspondence points using only local features. However, from a human perspective, matching is not solely based on local features; we also consider global features and semantics. For example, when matching vessels in two views, we first roughly align the branches before refining the details. This approach inspired us to introduce global information and semantics to guide the matching process. Recently, large image foundation models like DINOv2 \cite{oquab2023dinov2} have demonstrated the ability to capture semantic information, leading us to leverage these foundation models to enhance the matching process.

Specifically, We sample the global descriptor $g$ for each keypoint by interpolating the feature maps of foundation models. For global features $g_{A_i}$ and $g_{B_j}$, we compute the similarity matrix $G \in \mathbb{R}^{M \times N}$ between the points of two images(\Cref{fig:fig2}):

\begin{equation}
G_{ij} = \frac{g_{A_i} \cdot g_{B_j}}{|g_{A_i}| |g_{B_j}|}
\end{equation}

We then construct a mask $M$ based on the similarity matrix $G$. For each row $i$ in $G$, we set $M_{ij} = 1$ if $G_{ij}$ is within the top $k\%$ of the highest similarity values in row $i$, and $M_{ij} = 0$ otherwise:

\begin{equation}
M_{ij} = 
\begin{cases} 
1 & \text{if } G_{ij} \geq \text{threshold}(G_i, k) \\
0 & \text{otherwise}
\end{cases}
\end{equation}

\noindent where $\text{threshold}(G_i, k)$ represents the value above which $k\%$ of the highest similarity scores in row $i$ of $G$ are retained. 

This mask is applied to cross-attention layers to guide the feature aggregation between two images:

\begin{equation}
m_{i_\text{cross}} = \sum_{j \in S} \mathop{\text{Softmax}}_{\substack{k \in S}} \left( a_{ik}M_{ik} \right)_j v_j
\end{equation}

By incorporating this mask, it effectively filters out the features of points that might appear locally similar but, from a global perspective, should not be matched. This approach also guides the cross-attention process to focus on the most relevant and semantically appropriate points and mitigates the influence of misleading local similarities. 

\section{Experiments}

\subsection{Dataset}

\subsubsection{Synthetic Dataset}

We adopt the large-scale ImageCAS dataset \cite{zeng2023imagecas}, which contains 1,000 3D Computed Tomography Angiography (CTA) images along with segmentations. We reconstruct 3D meshes for the left coronary artery (LCA) and right coronary artery (RCA) based on the segmentation masks. Following the projection angles recommended in \cite{xu2017diagnostic}, we perturb these angles by intervals of \(5^{\circ}\) for rendering. This process yields 19 2D masks for the left anterior descending coronary artery (LAD), 24 masks for the left circumflex coronary artery (LCX), and 21 masks for the RCA. By pairing the images within each group, we generate 350 pairs (242 pairs for LCA, 108 pairs for RCA) per subject. The rendered masks have a resolution of \(512 \times 512\) pixels with a spatial resolution of 0.3 $mm$ per pixel. These masks are then fed into the diffusion model to generate the corresponding images. For all 1,000 subjects, we allocate 900 for training, 50 for validation, and 50 for testing, resulting in 315,000 pairs for training and 17,500 pairs each for validation and testing, respectively.

\subsubsection{Real Dataset}

We also construct an angiography image dataset, consisting of 21 real subjects and 47 pairs (27 pairs for LCA and 20 pairs for RCA). These images were collected from patients who underwent coronary artery-related interventions. The images are preprocessed by rescaling to a resolution of \(512 \times 512\) pixels with a pixel spacing of 0.308 mm. We then segment the preprocessed images to obtain the segmentation masks. The skeleton points of these masks are considered the centerline points of the vessels and are extracted for follow-up correspondence matching.

\subsection{Implementation Details}

The IDDPM model for generating angiography images was optimized using the AdamW optimizer with a learning rate of $10^{-4}$. It was trained for 200 epochs with 256 time steps. For correspondence matching, our model comprises 9 feature aggregation blocks and all descriptors have a feature dimension of 256. The foundational model guiding the training is DINOv2 ViT-14-base with registers. Our training of matching model consists of three stages. In the first two stages, we follow the training scheme from \cite{lindenberger2023lightglue}, initially pretraining on a Synthetic Homography dataset, which samples images from the Oxford and Paris datasets \cite{radenovic2018revisiting} and applies homography transformations to generate image pairs. The pretrained model is then further finetuned on the large-scale outdoor image dataset MegaDepth \cite{li2018megadepth}. More details refer to \cite{lindenberger2023lightglue}. Finally, we finetune the model on our synthetic dataset. The training epochs for each stage are 40/30/30, respectively, with a learning rate of $10^{-4}$. The training was performed with a batch size of 128 on 8 NVIDIA A100 GPUs.

\subsection{Qualitative Analysis of the Synthetic Data}

\begin{figure}[htb]
\begin{center}
\includegraphics[width=1\linewidth]{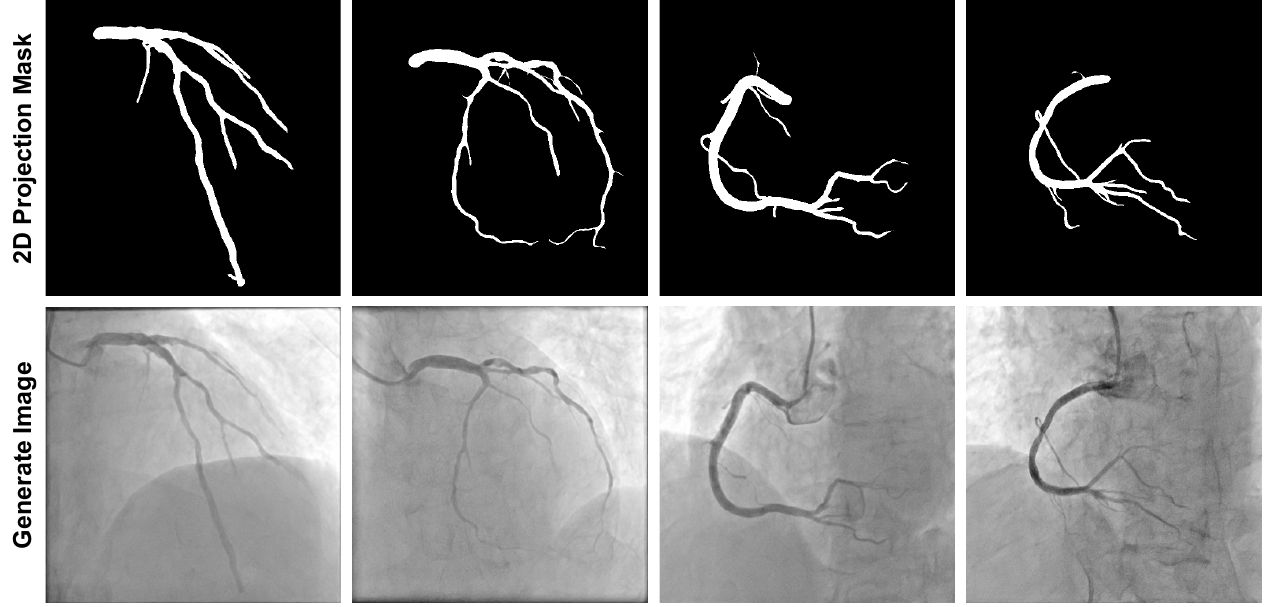}
\end{center}
\caption{Illustration of the masks and generated images from four randomly selected cases. The first row shows masks projected from the 3D mesh for the two LCA and two RCA cases, which are the condition for IDDPM model. The second row displays images generated by the trained IDDPM model.}
\label{fig:fig3}
\end{figure}

In this subsection, we qualitatively analyze the generation of synthetic data. In \Cref{fig:fig3}, we randomly select two cases for the LCA and RCA. The first row shows masks projected from the 3D mesh, while the second row displays images generated by the trained IDDPM model. Overall, the generated images exhibit high quality, with realistic backgrounds and vessel contrast. The branches in the masks are all preserved in the generated images. However, compared to real images, the generated images lack some specific details, such as very small vessel branches, due to the resolution limitations of CTA images. Typically, CTA has lower resolution compared to DSA, which results in missing tiny structures like small vessels in both segmentation and reconstructed mesh. Our synthetic data generation is conditioned on the projection masks of the 3D mesh which also lack these details. So in the generated image, those small vessels are missing. Notably, even without specific labels for the catheter and its tip, the diffusion model learned to add the catheter at the inlet of the coronary artery and simulate the effect of contrast injection, enhancing the realism. This is because angiography imaging inherently includes the catheter for contrast injection, and the diffusion model implicitly learned this from the training dataset.

\subsection{Matching Results on Synthetic Dataset}
\label{sec:syn}

In this subsection, we compare the proposed method with other baselines on the synthetic dataset to validate its effectiveness.  

\subsubsection{Baselines}

The compared baselines include: 1) Descriptors from SuperPoint \cite{detone2018superpoint} with the Mutual Nearest Neighbor (MNN) matching method. 2) SuperGlue \cite{sarlin2020superglue}, which utilizes descriptors from SuperPoint and employs self-attention and cross-attention for keypoint feature aggregation. 3) LightGlue \cite{lindenberger2023lightglue}, an improved version of SuperGlue, offering better performance and speed. 4) LightGlue with foundation model (DINOv2) guidance, where we add the mask generated by the foundation model to the cross-attention block of LightGlue. 5) The proposed method, incorporating foundation guidance and training on the synthetic dataset.

\subsubsection{Evaluation Metric}
On the synthetic dataset, since we have the ground truth labels for correct matches, we compute the area under the cumulative error curve (AUC) for matches within 1px and 3px of the ground truth points. We also estimate the pose of the projection camera from robust correspondences using RANSAC\cite{fischler1981random}, and report the pose accuracy (percentage of correct poses within \(15^{\circ}\) and \(30^{\circ}\) of error) and the AUC.

\subsubsection{Results}

\begin{table}[t]
\centering
\caption{Matching results on the synthetic dataset. The table compares different methods based on Pose AUC at $15^{\circ}$ and $30^{\circ}$, as well as Match AUC at 1px and 3px.}
\begin{tabular}{lccccc}
\hline
\multirow{2}{*}{Method} & \multicolumn{2}{c}{Pose AUC} & \multicolumn{2}{c}{Match AUC} \\
 & @ $15^{\circ}$ & @ $30^{\circ}$ & @ 1px & @ 3px \\
\hline
Superpoint+NN & 0.0147 & 0.1185 & 0.0989 & 0.1180 \\
SuperGlue & 0.0631 & 0.1841 & 0.2315 & 0.2647 \\
LightGlue & 0.0485 & 0.1563 & 0.2904 & 0.3312 \\
LightGlue w/ Guidance & 0.0576 & 0.1745 & 0.2951 & 0.3367 \\
Ours & \textbf{0.8154} & \textbf{0.8777} & \textbf{0.8692} & \textbf{0.8868} \\
\hline
\end{tabular}
\label{tab:synthetic}
\end{table}

\Cref{tab:synthetic} presents the matching results on the synthetic dataset, comparing the proposed method against several baselines. The proposed method outperforms all baselines, achieving a Pose AUC of 0.8154 at $15^{\circ}$ and 0.8777 at $30^{\circ}$. For Match AUC at 1px and 3px, it achieves scores of 0.8692 and 0.8868, respectively, demonstrating superior matching accuracy. In contrast, Superpoint+NN performs the worst compared to the learning-based methods. Both SuperGlue and LightGlue struggle to generalize well to the synthetic dataset. However, applying the similarity mask from DINOv2 on LightGlue improves matching performance, making it the closest baseline to the proposed method and indicating the effectiveness of foundation model guidance.

\subsection{Matching Results on Real Dataset}

\begin{figure*}[htb]
\begin{center}
\includegraphics[width=1\linewidth]{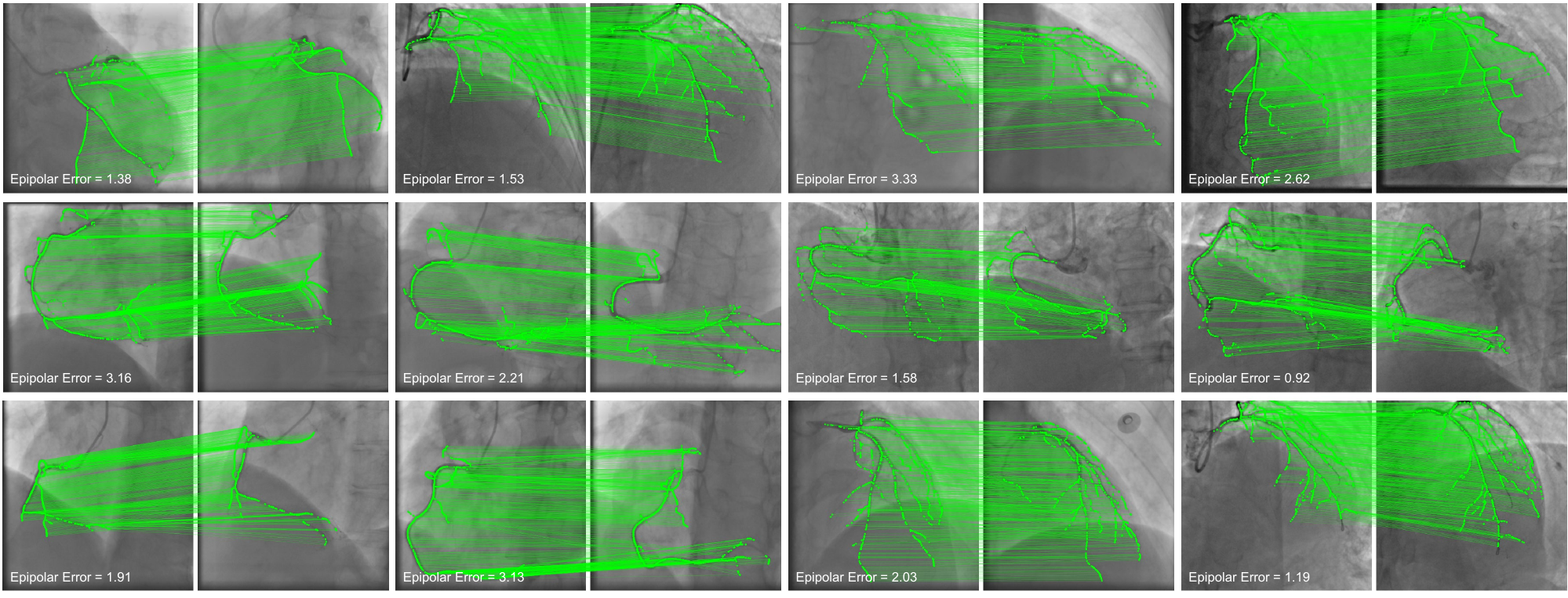}
\end{center}
\caption{Visualization of the matched keypoints using our method from 12 randomly selected pairs in real dataset. The matched keypoints are denoted in green dots, with correspondences indicated by green lines. The mean epipolar error for each image pair is shown below each sub-figure.}
\label{fig:fig4}
\end{figure*}

\subsubsection{Baselines}
We adopted the same baselines as in \Cref{sec:syn}. Additionally, we ablated the foundation model guidance in our proposed model, maintaining the same training scheme as another baseline to validate the effectiveness of the foundation model guidance.

\subsubsection{Evaluation Metric}
On the real dataset, ground truth matches are unavailable. We estimate the pose of the projection camera using robust correspondences with RANSAC \cite{fischler1981random}. Due to cardiac motion and respiratory effects, we focus only on validating the estimation of the rotation vector, as the translation vectors serve as extrinsic parameters for calibration. Additionally, using the estimated pose, we compute the mean and standard deviation of the epipolar errors of the matches from each method. 

\subsubsection{Results}

\begin{table}[h]
\centering
\caption{Matching results on the real dataset, comparing different methods in terms of Pose AUC at $15^{\circ}$ and $30^{\circ}$, as well as Epipolar Error (Mean $\pm$ Std).}

\begin{tabular}{lccc}
\hline
\multirow{2}{*}{Method} & \multicolumn{2}{c}{Pose AUC} & Epipolar Error \\
 & @ $15^{\circ}$ & @ $30^{\circ}$ & (Mean $\pm$ Std) \\
\hline
Superpoint+NN & 0.0106 & 0.0644 & 50.70 $\pm$ 19.14 \\
SuperGlue & 0 & 0.0730 & 18.62 $\pm$ 18.50 \\
LightGlue & 0 & 0.0525 & 5.73 $\pm$ 5.66 \\
LightGlue w/ Guidance & 0 & 0.0755 & 5.34 $\pm$ 3.81 \\
Ours w/o Guidance & 0.0193 & 0.1157 & 4.69 $\pm$ 5.35 \\
Ours & \textbf{0.0382} & \textbf{0.1624} & \textbf{2.57} $\pm$ \textbf{1.44} \\
\hline
\end{tabular}
\label{tab:real}
\end{table}

\Cref{tab:real} presents the results on the real dataset. The AUC for pose estimation is relatively low because X-ray images, as transmission images, differ significantly from natural images where keypoints lie on a surface plane. In X-ray images, even slight movements of a keypoint in 2D space can significantly impact its position in 3D space. For the synthetic dataset, we use the ground truth centerline points projected from the 3D space as keypoints. In contrast, for the real dataset, the keypoints are centerline points that may not correspond to the true centerline points in 3D space. This discrepancy results in a lower pose estimation AUC. Despite this, for pose estimation within $30^{\circ}$, our method outperforms the closest baseline by a substantial margin, achieving a Pose AUC of 0.1624 compared to 0.1157 without foundation model guidance.

The epipolar error measures the distance between matched points and their corresponding epipolar lines. A smaller epipolar error indicates more accurate matches, as correct correspondences result in points lying closer to the epipolar lines. In \Cref{tab:real}, it is observed that SuperGlue doesn't generalize as well as LightGlue. Our method achieves the lowest mean epipolar error (2.57) and standard deviation (1.44), significantly better than the closest baseline. This indicates that the majority of matches produced by our method are accurate, leading to improved robustness in pose estimation. By applying the mask from DINOv2, the epipolar error of the LightGlue method also decreases, consistent with the results on the synthetic dataset. However, when ablating the foundation model guidance, our model's performance drops significantly, indicating that the foundation model is crucial in improving the generalizability of the model. 

In \Cref{fig:fig4}, we visualize the matched keypoints of our method. Most centerline points are correctly matched with the points at the corresponding branch, demonstrating the accuracy of our approach. However, for the RCA which has a higher epipolar error, some side branches are mistakenly matched to incorrect branches. This issue arises because, compared to the LCA, the RCA has fewer side branches and bifurcations, and is with significant occlusion in some areas. Additionally, the matching process relies on keypoints derived from the segmentation, and some side branches may not be fully segmented, resulting in potential mismatches.

Overall, these results highlight the effectiveness of our method in accurately matching keypoints in complex vascular structures.

\begin{figure*}[htb]
\begin{center}
\includegraphics[width=1\linewidth]{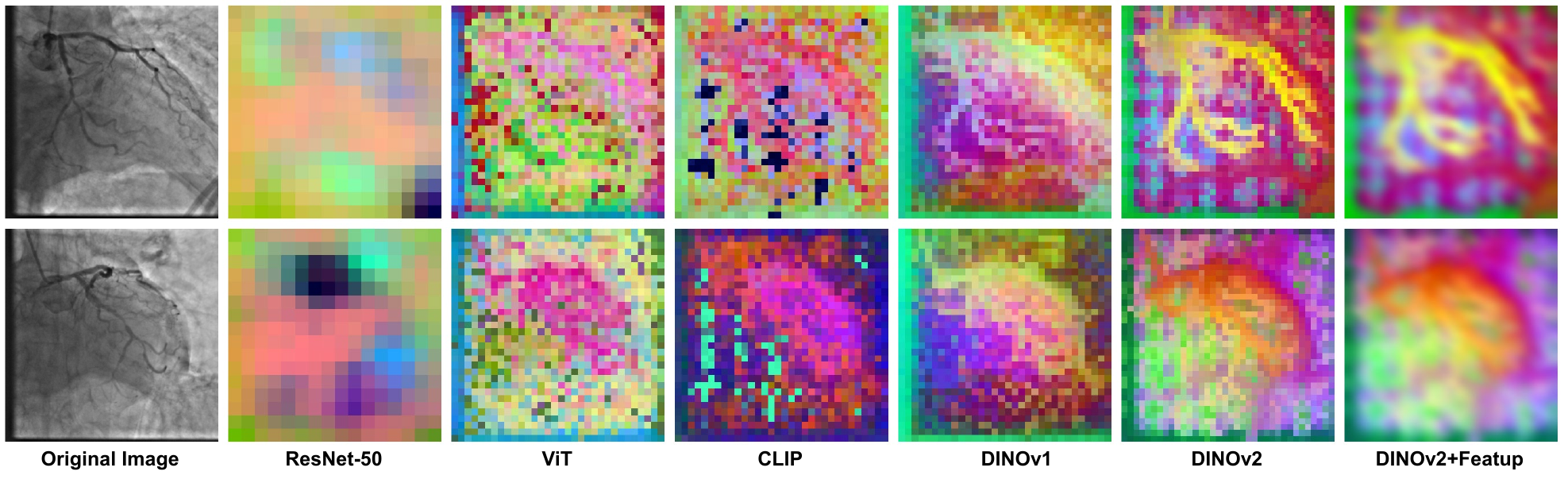}
\end{center}
\caption{Visualization of 3-dimensional PCA applied to the features of various foundation models.}
\label{fig:fig5}
\end{figure*}

\subsection{Efficacy of Foundation Models}
In this subsection, we explore the efficacy of different foundation models in improving the generalizability of matching methods. We use our method without any foundation model guidance as the baseline and apply the similarity masks generated by the features of various foundation models to the cross-attention block. We compare features generated by ResNet \cite{he2016deep}, Vision Transformer (ViT) \cite{dosovitskiy2020image}, Contrastive Language-Image Pre-Training (CLIP) \cite{radford2021learning}, DINOv1 \cite{caron2021emerging}, DINOv2 \cite{oquab2023dinov2,darcet2023vitneedreg}, and DINOv2 with Featup \cite{fu2024featup} which improve the spatial resolution of the DINOv2's feature map. The results are shown in \Cref{tab:foundation_models}.

\begin{table}[h]
\centering
\caption{Performance comparison of foundation models on pose estimation and epipolar error for real dataset.}
\begin{tabular}{lcccc}
\hline
\multirow{2}{*}{Foundation Models} & \multicolumn{2}{c}{Pose AUC} & Epipolar Error \\
 & @ $15^{\circ}$ & @ $30^{\circ}$ & (Mean $\pm$ Std) \\
\hline
Baseline (w/o Guidance) & 0.0193 & 0.1157 & 4.69 $\pm$ 5.35 \\
ResNet-50 & 0.0208 & 0.1121 & 4.68 $\pm$ 3.97 \\
ViT & 0.0381 & 0.1141 & 3.81 $\pm$ 4.65 \\
CLIP & \textbf{0.0687} & 0.1453 & 3.46 $\pm$ 3.39 \\
DINOv1 & 0.0358 & 0.1237 & 3.25 $\pm$ 2.87 \\
DINOv2 & 0.0372 & 0.1458 & 2.79 $\pm$ 1.84 \\
DINOv2+Featup & 0.0382 & \textbf{0.1624} & \textbf{2.57} $\pm$ \textbf{1.44} \\
\hline
\end{tabular}
\label{tab:foundation_models}
\end{table}

It is observed that our baseline model, without any foundation model guidance, demonstrates limited performance. ResNet-50, despite being a widely used model, does not perform well in this context, showing minimal improvement over the baseline. This suggests that ResNet-50's features pretrained on ImageNet dataset may not be generalizable for other vision task. In contrast, models like DINOv2, significantly enhance performance, effectively improving the model’s ability to accurately match keypoints and reducing errors. DINOv2+Featup demonstrates the superior performance among all compared method, indicating the importance of fine-grained features in achieving better generalizability and precision. Other models, such as CLIP and DINOv1, which were also trained on large scale dataset, also show notable improvements. 

To further explore the differences in how foundation models guide correspondence matching, we follow the method in \cite{oquab2023dinov2,fu2024featup} by applying a 3-dimensional PCA to the features of each foundation model and visualized them in \Cref{fig:fig5}. For models with limited performance, such as ResNet-50, we did not observe any meaningful patterns. In contrast, for DINOv2, the visualization clearly distinguishes the vessels from the background in different colors, indicating more discernible and meaningful features. DINOv2 shows more distinct and accurate feature representation than DINOv1 and CLIP, which correlates with its higher performance in correspondence matching. Featup enhances the resolution of the DINOv2 features, leading to the highest performance among all models. These results highlights the importance of semantic-rich features for guiding the correspondence matching, as better semantic understanding guides the model to more focus on the semantically corresponding branches and point, improving overall matching accuracy.

These results collectively demonstrate the critical role of foundation models in advancing the performance of matching methods in our task.

\section{Conclusion}
Our proposed pipeline effectively addresses the challenges of correspondence matching in coronary angiography by utilizing diffusion models to generate realistic synthetic datasets. By incorporating large-scale image foundation models, our approach significantly enhances matching accuracy by focusing on semantically relevant regions and keypoints. This method demonstrates superior performance on both synthetic and real datasets, offering a robust and practical solution for camera calibration in 3D reconstruction of cornary artery. Furthermore, our exploration of various foundation models provides valuable insights into their applicability in medical imaging, paving the way for future advancements in medical imaging applications.

\bibliographystyle{IEEEtran.bst}
\bibliography{egbib}

\begin{thebibliography}{10}
\providecommand{\url}[1]{#1}
\csname url@samestyle\endcsname
\providecommand{\newblock}{\relax}
\providecommand{\bibinfo}[2]{#2}
\providecommand{\BIBentrySTDinterwordspacing}{\spaceskip=0pt\relax}
\providecommand{\BIBentryALTinterwordstretchfactor}{4}
\providecommand{\BIBentryALTinterwordspacing}{\spaceskip=\fontdimen2\font plus
\BIBentryALTinterwordstretchfactor\fontdimen3\font minus \fontdimen4\font\relax}
\providecommand{\BIBforeignlanguage}[2]{{%
\expandafter\ifx\csname l@#1\endcsname\relax
\typeout{** WARNING: IEEEtran.bst: No hyphenation pattern has been}%
\typeout{** loaded for the language `#1'. Using the pattern for}%
\typeout{** the default language instead.}%
\else
\language=\csname l@#1\endcsname
\fi
#2}}
\providecommand{\BIBdecl}{\relax}
\BIBdecl

\bibitem{go2014heart}
A.~S. Go, D.~Mozaffarian, V.~L. Roger, E.~J. Benjamin, J.~D. Berry, M.~J. Blaha, S.~Dai, E.~S. Ford, C.~S. Fox, S.~Franco \emph{et~al.}, ``Heart disease and stroke statistics—2014 update: a report from the american heart association,'' \emph{circulation}, vol. 129, no.~3, pp. e28--e292, 2014.

\bibitem{townsend2016cardiovascular}
N.~Townsend, L.~Wilson, P.~Bhatnagar, K.~Wickramasinghe, M.~Rayner, and M.~Nichols, ``Cardiovascular disease in europe: epidemiological update 2016,'' \emph{European heart journal}, vol.~37, no.~42, pp. 3232--3245, 2016.

\bibitem{de2008fractional}
B.~De~Bruyne and J.~Sarma, ``Fractional flow reserve: a review,'' \emph{Heart}, vol.~94, no.~7, pp. 949--959, 2008.

\bibitem{green2004three}
N.~E. Green, S.-Y.~J. Chen, J.~C. Messenger, B.~M. Groves, and J.~D. Carroll, ``Three-dimensional vascular angiography,'' \emph{Current problems in cardiology}, vol.~29, no.~3, pp. 104--142, 2004.

\bibitem{katritsis1991limitations}
D.~Katritsis and M.~Webb-Peploe, ``Limitations of coronary angiography: an underestimated problem?'' \emph{Clinical cardiology}, vol.~14, no.~1, pp. 20--24, 1991.

\bibitem{gollapudi2007utility}
R.~R. Gollapudi, R.~Valencia, S.~S. Lee, G.~B. Wong, P.~S. Teirstein, and M.~J. Price, ``Utility of three-dimensional reconstruction of coronary angiography to guide percutaneous coronary intervention,'' \emph{Catheterization and Cardiovascular Interventions}, vol.~69, no.~4, pp. 479--482, 2007.

\bibitem{yang2009novel}
J.~Yang, Y.~Wang, Y.~Liu, S.~Tang, and W.~Chen, ``Novel approach for 3-d reconstruction of coronary arteries from two uncalibrated angiographic images,'' \emph{Ieee transactions on image processing}, vol.~18, no.~7, pp. 1563--1572, 2009.

\bibitem{bourantas2005method}
C.~V. Bourantas, I.~C. Kourtis, M.~E. Plissiti, D.~I. Fotiadis, C.~S. Katsouras, M.~I. Papafaklis, and L.~K. Michalis, ``A method for 3d reconstruction of coronary arteries using biplane angiography and intravascular ultrasound images,'' \emph{Computerized Medical Imaging and Graphics}, vol.~29, no.~8, pp. 597--606, 2005.

\bibitem{vukicevic2018three}
A.~M. Vukicevic, S.~{\c{C}}imen, N.~Jagic, G.~Jovicic, A.~F. Frangi, and N.~Filipovic, ``Three-dimensional reconstruction and nurbs-based structured meshing of coronary arteries from the conventional x-ray angiography projection images,'' \emph{Scientific reports}, vol.~8, no.~1, p. 1711, 2018.

\bibitem{ccimen2016reconstruction}
S.~{\c{C}}imen, A.~Gooya, M.~Grass, and A.~F. Frangi, ``Reconstruction of coronary arteries from x-ray angiography: A review,'' \emph{Medical image analysis}, vol.~32, pp. 46--68, 2016.

\bibitem{baka2015respiratory}
N.~Baka, B.~Lelieveldt, C.~Schultz, W.~Niessen, and T.~van Walsum, ``Respiratory motion estimation in x-ray angiography for improved guidance during coronary interventions,'' \emph{Physics in Medicine \& Biology}, vol.~60, no.~9, p. 3617, 2015.

\bibitem{blondel2006reconstruction}
C.~Blondel, G.~Malandain, R.~Vaillant, and N.~Ayache, ``Reconstruction of coronary arteries from a single rotational x-ray projection sequence,'' \emph{IEEE Transactions on Medical Imaging}, vol.~25, no.~5, pp. 653--663, 2006.

\bibitem{sarlin2020superglue}
P.-E. Sarlin, D.~DeTone, T.~Malisiewicz, and A.~Rabinovich, ``Superglue: Learning feature matching with graph neural networks,'' in \emph{Proceedings of the IEEE/CVF conference on computer vision and pattern recognition}, 2020, pp. 4938--4947.

\bibitem{lindenberger2023lightglue}
P.~Lindenberger, P.-E. Sarlin, and M.~Pollefeys, ``Lightglue: Local feature matching at light speed,'' in \emph{Proceedings of the IEEE/CVF International Conference on Computer Vision}, 2023, pp. 17\,627--17\,638.

\bibitem{xu2017diagnostic}
B.~Xu, S.~Tu, S.~Qiao, X.~Qu, Y.~Chen, J.~Yang, L.~Guo, Z.~Sun, Z.~Li, F.~Tian \emph{et~al.}, ``Diagnostic accuracy of angiography-based quantitative flow ratio measurements for online assessment of coronary stenosis,'' \emph{Journal of the American College of Cardiology}, vol.~70, no.~25, pp. 3077--3087, 2017.

\bibitem{li2018megadepth}
Z.~Li and N.~Snavely, ``Megadepth: Learning single-view depth prediction from internet photos,'' in \emph{Proceedings of the IEEE conference on computer vision and pattern recognition}, 2018, pp. 2041--2050.

\bibitem{oquab2023dinov2}
M.~Oquab, T.~Darcet, T.~Moutakanni, H.~Vo, M.~Szafraniec, V.~Khalidov, P.~Fernandez, D.~Haziza, F.~Massa, A.~El-Nouby \emph{et~al.}, ``Dinov2: Learning robust visual features without supervision,'' \emph{arXiv preprint arXiv:2304.07193}, 2023.

\bibitem{jiang2024omniglue}
H.~Jiang, A.~Karpur, B.~Cao, Q.~Huang, and A.~Araujo, ``Omniglue: Generalizable feature matching with foundation model guidance,'' in \emph{Proceedings of the IEEE/CVF Conference on Computer Vision and Pattern Recognition}, 2024, pp. 19\,865--19\,875.

\bibitem{liu2017deep}
Y.~Liu, J.~Pan, and Z.~Su, ``Deep feature matching for dense correspondence,'' in \emph{2017 IEEE International Conference on Image Processing (ICIP)}.\hskip 1em plus 0.5em minus 0.4em\relax IEEE, 2017, pp. 795--799.

\bibitem{sun2021loftr}
J.~Sun, Z.~Shen, Y.~Wang, H.~Bao, and X.~Zhou, ``Loftr: Detector-free local feature matching with transformers,'' in \emph{Proceedings of the IEEE/CVF conference on computer vision and pattern recognition}, 2021, pp. 8922--8931.

\bibitem{cong2013energy}
W.~Cong, J.~Yang, Y.~Liu, and Y.~Wang, ``Energy back-projective composition for 3-d coronary artery reconstruction,'' in \emph{2013 35th Annual International Conference of the IEEE Engineering in Medicine and Biology Society (EMBC)}.\hskip 1em plus 0.5em minus 0.4em\relax IEEE, 2013, pp. 5151--5154.

\bibitem{cong2015quantitative}
W.~Cong, J.~Yang, D.~Ai, Y.~Chen, Y.~Liu, and Y.~Wang, ``Quantitative analysis of deformable model-based 3-d reconstruction of coronary artery from multiple angiograms,'' \emph{IEEE Transactions on Biomedical Engineering}, vol.~62, no.~8, pp. 2079--2090, 2015.

\bibitem{yang2014external}
J.~Yang, W.~Cong, Y.~Chen, J.~Fan, Y.~Liu, and Y.~Wang, ``External force back-projective composition and globally deformable optimization for 3-d coronary artery reconstruction,'' \emph{Physics in Medicine \& Biology}, vol.~59, no.~4, p. 975, 2014.

\bibitem{kunio2018vessel}
M.~Kunio, C.~C. O’Brien, A.~C. Lopes, L.~Bailey, P.~A. Lemos, G.~J. Tearney, and E.~R. Edelman, ``Vessel centerline reconstruction from non-isocentric and non-orthogonal paired monoplane angiographic images,'' \emph{The international journal of cardiovascular imaging}, vol.~34, pp. 673--682, 2018.

\bibitem{jiang2021cotr}
W.~Jiang, E.~Trulls, J.~Hosang, A.~Tagliasacchi, and K.~M. Yi, ``Cotr: Correspondence transformer for matching across images,'' in \emph{Proceedings of the IEEE/CVF International Conference on Computer Vision}, 2021, pp. 6207--6217.

\bibitem{ho2020denoising}
J.~Ho, A.~Jain, and P.~Abbeel, ``Denoising diffusion probabilistic models,'' \emph{Advances in neural information processing systems}, vol.~33, pp. 6840--6851, 2020.

\bibitem{song2020score}
Y.~Song, J.~Sohl-Dickstein, D.~P. Kingma, A.~Kumar, S.~Ermon, and B.~Poole, ``Score-based generative modeling through stochastic differential equations,'' \emph{arXiv preprint arXiv:2011.13456}, 2020.

\bibitem{song2020denoising}
J.~Song, C.~Meng, and S.~Ermon, ``Denoising diffusion implicit models,'' \emph{arXiv preprint arXiv:2010.02502}, 2020.

\bibitem{wu2024medsegdiff}
J.~Wu, R.~Fu, H.~Fang, Y.~Zhang, Y.~Yang, H.~Xiong, H.~Liu, and Y.~Xu, ``Medsegdiff: Medical image segmentation with diffusion probabilistic model,'' in \emph{Medical Imaging with Deep Learning}.\hskip 1em plus 0.5em minus 0.4em\relax PMLR, 2024, pp. 1623--1639.

\bibitem{chen2024federated}
X.~Chen, S.~Zhang, E.~Z. Chen, Y.~Liu, L.~Zhao, T.~Chen, and S.~Sun, ``Federated data model,'' \emph{arXiv preprint arXiv:2403.08887}, 2024.

\bibitem{qiu2024clinically}
S.~Qiu, S.~Pan, Y.~Liu, L.~Zhao, J.~Xu, Q.~Liu, T.~Chen, E.~Z. Chen, X.~Chen, and S.~Sun, ``Clinically feasible diffusion reconstruction for highly-accelerated cardiac cine mri,'' \emph{arXiv preprint arXiv:2403.08749}, 2024.

\bibitem{qiu2024spatiotemporal}
------, ``Spatiotemporal diffusion model with paired sampling for accelerated cardiac cine mri,'' \emph{arXiv preprint arXiv:2403.08758}, 2024.

\bibitem{nichol2021improved}
A.~Q. Nichol and P.~Dhariwal, ``Improved denoising diffusion probabilistic models,'' in \emph{International conference on machine learning}.\hskip 1em plus 0.5em minus 0.4em\relax PMLR, 2021, pp. 8162--8171.

\bibitem{dosovitskiy2020image}
A.~Dosovitskiy, L.~Beyer, A.~Kolesnikov, D.~Weissenborn, X.~Zhai, T.~Unterthiner, M.~Dehghani, M.~Minderer, G.~Heigold, S.~Gelly \emph{et~al.}, ``An image is worth 16x16 words: Transformers for image recognition at scale,'' \emph{arXiv preprint arXiv:2010.11929}, 2020.

\bibitem{radford2021learning}
A.~Radford, J.~W. Kim, C.~Hallacy, A.~Ramesh, G.~Goh, S.~Agarwal, G.~Sastry, A.~Askell, P.~Mishkin, J.~Clark \emph{et~al.}, ``Learning transferable visual models from natural language supervision,'' in \emph{International conference on machine learning}.\hskip 1em plus 0.5em minus 0.4em\relax PMLR, 2021, pp. 8748--8763.

\bibitem{he2022masked}
K.~He, X.~Chen, S.~Xie, Y.~Li, P.~Doll{\'a}r, and R.~Girshick, ``Masked autoencoders are scalable vision learners,'' in \emph{Proceedings of the IEEE/CVF conference on computer vision and pattern recognition}, 2022, pp. 16\,000--16\,009.

\bibitem{caron2021emerging}
M.~Caron, H.~Touvron, I.~Misra, H.~J\'egou, J.~Mairal, P.~Bojanowski, and A.~Joulin, ``Emerging properties in self-supervised vision transformers,'' in \emph{Proceedings of the International Conference on Computer Vision (ICCV)}, 2021.

\bibitem{detone2018superpoint}
D.~DeTone, T.~Malisiewicz, and A.~Rabinovich, ``Superpoint: Self-supervised interest point detection and description,'' in \emph{Proceedings of the IEEE conference on computer vision and pattern recognition workshops}, 2018, pp. 224--236.

\bibitem{zeng2023imagecas}
A.~Zeng, C.~Wu, G.~Lin, W.~Xie, J.~Hong, M.~Huang, J.~Zhuang, S.~Bi, D.~Pan, N.~Ullah \emph{et~al.}, ``Imagecas: A large-scale dataset and benchmark for coronary artery segmentation based on computed tomography angiography images,'' \emph{Computerized Medical Imaging and Graphics}, vol. 109, p. 102287, 2023.

\bibitem{radenovic2018revisiting}
F.~Radenovi{\'c}, A.~Iscen, G.~Tolias, Y.~Avrithis, and O.~Chum, ``Revisiting oxford and paris: Large-scale image retrieval benchmarking,'' in \emph{Proceedings of the IEEE conference on computer vision and pattern recognition}, 2018, pp. 5706--5715.

\bibitem{fischler1981random}
M.~A. Fischler and R.~C. Bolles, ``Random sample consensus: a paradigm for model fitting with applications to image analysis and automated cartography,'' \emph{Communications of the ACM}, vol.~24, no.~6, pp. 381--395, 1981.

\bibitem{he2016deep}
K.~He, X.~Zhang, S.~Ren, and J.~Sun, ``Deep residual learning for image recognition,'' in \emph{Proceedings of the IEEE conference on computer vision and pattern recognition}, 2016, pp. 770--778.

\bibitem{darcet2023vitneedreg}
T.~Darcet, M.~Oquab, J.~Mairal, and P.~Bojanowski, ``Vision transformers need registers,'' 2023.

\bibitem{fu2024featup}
\BIBentryALTinterwordspacing
S.~Fu, M.~Hamilton, L.~E. Brandt, A.~Feldmann, Z.~Zhang, and W.~T. Freeman, ``Featup: A model-agnostic framework for features at any resolution,'' in \emph{The Twelfth International Conference on Learning Representations}, 2024. [Online]. Available: \url{https://openreview.net/forum?id=GkJiNn2QDF}
\BIBentrySTDinterwordspacing

\end{thebibliography}

\end{document}